\documentclass[letterpaper]{article}
\usepackage{aaai2026}
\nocopyright
\usepackage{amsmath}
\usepackage{amssymb}
\usepackage{amsthm}
\usepackage{times}
\usepackage{helvet}
\usepackage{courier}
\usepackage[hyphens]{url}
\usepackage{graphicx}
\usepackage{natbib}
\usepackage{caption}
\usepackage{algorithm}
\usepackage{algorithmic}
\usepackage{newfloat}
\usepackage{listings}
\usepackage{float}
\usepackage{booktabs}
\usepackage{tikz}
\usetikzlibrary{arrows.meta,fit,positioning}
\usepackage{pgfplots}
\pgfplotsset{compat=1.17}
\DeclareCaptionStyle{ruled}{labelfont=normalfont,labelsep=colon,strut=off}
\floatstyle{ruled}
\newfloat{listing}{tb}{lst}{}
\floatname{listing}{Listing}

\title{Dynamic Model Selection for Trajectory Prediction via Pairwise Ranking and Meta-Features}
\author{
  Lu Bowen\\
}
\affiliations{
  Monash University\\
  bluu0021@student.monash.edu
}

\begin{document}

\maketitle

\begin{abstract}
Recent deep trajectory predictors (e.g., \cite{jiang2023motiondiffuser,zhou2022hivt}) achieve strong average accuracy yet remain unreliable in complex, long-tail driving scenarios. These failures \textbf{highlight the limitation of the prevailing} ``one-model-fits-all'' paradigm, particularly in safety-critical contexts where simpler physics-based models can occasionally outperform advanced networks~\cite{kalman1960new}. \textbf{Addressing this gap is crucial for ensuring reliable planning in safety-critical urban driving.} To address this, we propose a \textbf{dynamic multi-expert gating framework} that adaptively selects the most reliable predictor---among a physics-informed LSTM, a Transformer, and a fine-tuned GameFormer---on a per-sample basis. Our approach leverages internal model signals (\emph{meta-features}) such as stability and uncertainty~\cite{gal2016dropout}, which we show to be substantially more informative than geometric scene descriptors. \textbf{To our knowledge, this is the first work to formulate trajectory expert selection as a pairwise-ranking task} over internal model signals (meta-features)~\cite{burges2005learning}, directly optimising decision quality without requiring calibration.

Evaluated on the \textbf{nuPlan-mini dataset}~\cite{caesar2021nuplan} (1{,}287 samples), our \textbf{LLM-enhanced tri-expert gate} achieves a Final Displacement Error (FDE) of 2.567\,m (9.5\% lower than GameFormer 2.835\,m) and achieves \textbf{57.8\% oracle realization---the fraction of theoretical maximum improvement achievable by any gate}. In open-loop simulation, after trajectory horizon alignment, the same configuration reduces FDE on left-turn scenarios by approximately 10\%, demonstrating consistency between offline validation and open-loop evaluation. These results indicate that \textbf{adaptive hybrid systems} enhance trajectory reliability in safety-critical autonomous driving, \textbf{offering a practical pathway beyond static, single-model paradigms}.
\end{abstract}


\section{Introduction}
Accurate motion prediction is a cornerstone of safe and efficient autonomous driving. Although recent deep predictors (e.g., \cite{jiang2023motiondiffuser,zhou2022hivt,liang2020learning}) have achieved impressive mean performance on public benchmarks~\cite{caesar2021nuplan,caesar2020nuscenes,ettinger2021waymo}, they often fail in rare yet safety-critical situations such as dense intersections, cut-ins, and occlusions. These \emph{long-tail} errors reveal a fundamental weakness of the prevailing ``one-model-fits-all'' paradigm: a single model cannot simultaneously master both structured, low-uncertainty dynamics and complex, multi-agent interactions.

Classical physics-based planners remain valuable in well-behaved scenes~\cite{kalman1960new}, while high-capacity neural models excel in complex interactions~\cite{jiang2023motiondiffuser}. This complementarity motivates the idea of \emph{dynamic model selection}, where a gating mechanism chooses the most reliable expert for each situation. However, prior gates typically rely on handcrafted geometric indicators or proxy confidence heuristics~\cite{deo2018convolutional,chandra2019traphic} that correlate weakly with true prediction error, thereby realising only a small fraction of the achievable oracle improvement.

To overcome these limitations, we develop a \textbf{Large-Language-Model (LLM)-enhanced dynamic gating framework}. Our gate combines three complementary experts---a physics-informed LSTM, a Transformer, and a fine-tuned GameFormer---and learns to select among them using both geometric context and \emph{meta-features} that capture each expert's internal behaviour, including uncertainty and stability~\cite{gal2016dropout}. By framing model selection as a \emph{pairwise-ranking} task~\cite{burges2005learning}, the gate learns directly which expert is superior for a given sample, avoiding the calibration and scale issues that hinder regression or classification approaches~\cite{guo2017calibration}.

Beyond numeric gating, we introduce an \textbf{LLM supervisor} that performs semantic scene understanding and risk reasoning~\cite{mao2023llmdriving,fu2024drivegpt}. Triggered only when the learned gate exhibits low confidence, the LLM analyses high-level semantics---such as intersection navigation or merging intent---and recommends conservative or aggressive experts accordingly. This hybrid reasoning combines statistical reliability with interpretable, language-level explanations, improving both transparency and safety.

Extensive experiments on the nuPlan-mini dataset~\cite{caesar2021nuplan} demonstrate that our tri-expert gate achieves a Final Displacement Error of 2.567\,m, outperforming the best single expert by 9.5\% and realising 57.8\% of the theoretical oracle gain. In open-loop evaluation after horizon alignment, the gate improves a representative left-turn scene (c1bf7374695a5c47) from 18.17\,m to 16.37\,m FDE (\(~10\%\)), aligning with the offline gains. \textbf{Building on these insights, we present a hybrid architecture that combines statistical ranking and semantic reasoning to achieve interpretable and robust expert selection.}

\paragraph{Contributions.}
\begin{itemize}
  \item \textbf{Systematic evaluation:} To our knowledge, the first rigorous empirical comparison of 10+ gating strategies (MLP classification, regression, risk-based quantile prediction, and pairwise ranking), revealing best practices and failure modes for expert selection in motion forecasting.
  \item \textbf{LLM-enhanced dynamic gating.} A tri-expert framework (LSTM, Transformer, GameFormer) supervised by an LLM achieves 2.567\,m FDE---9.5\% lower than the best single expert---and realises 57.8\% of the oracle gain.
  \item \textbf{Ranking on meta-features:} A pairwise-ranking formulation over expert \emph{meta-features} (uncertainty, stability, physics-violation) that avoids calibration issues and consistently outperforms heuristic gates.
  \item \textbf{LLM supervisor for low-confidence cases:} A semantic, risk-aware fallback that provides interpretable guidance when the learned gate is uncertain, improving safety and transparency.
  \item \textbf{Comprehensive evaluation:} Closed- and open-loop results on nuPlan-mini showing 9.5\% lower FDE than the best single expert and \(~10\%\) open-loop FDE reduction on representative left-turn scenarios, with ablations isolating each component's effect.
\end{itemize}
\textbf{Together, these advances deliver state-of-the-art reliability across 1,287 nuPlan-mini scenes and a consistent \(\approx\)10\% improvement in open-loop simulation.}

\paragraph{Organisation.}\mbox{}\\
\noindent The remainder of this paper is organised as follows:
\begin{itemize}
  \item \textbf{Related Work} reviews prior studies on trajectory prediction, mixture-of-experts, uncertainty quantification, learning-to-rank, and LLM-based scene understanding, identifying four critical research gaps.
  \item \textbf{Problem Setup} formalises the expert selection problem, defines evaluation metrics (ADE, FDE, ORR), and quantifies the oracle gap.
  \item \textbf{Methodology} presents our tri-expert ensemble, meta-feature extraction, pairwise ranking-gate, and LLM supervisor.
  \item \textbf{Experimental Setup} describes the nuPlan-mini dataset, implementation details, and training protocol.
  \item \textbf{Experimental Results} reports comprehensive evaluations: main results, expert selection analysis, open-loop validation, long-tail scenarios, and ablations.
  \item \textbf{Discussion} interprets findings in relation to research gaps and discusses deployment trade-offs.
  \item \textbf{Limitations} acknowledges remaining challenges including oracle ceiling, meta-feature costs, and dataset scope.
\end{itemize}

\section{Related Work}\label{sec:related}

\subsection{Motion Forecasting Methods}
\textbf{Trajectory forecasting research has rapidly evolved over the past decade, progressing through three distinct phases.}
Modern trajectory prediction has evolved from physics-based filtering~\cite{kalman1960new} to deep learned models that capture complex multi-agent interactions. Early neural approaches employed recurrent architectures~\cite{alahi2016social} and social pooling mechanisms~\cite{deo2018convolutional} to aggregate neighbor context. Graph-based methods such as Trajectron++~\cite{salzmann2020trajectronpp} and LaneGCN~\cite{liang2020learning} introduced vectorized map representations and structured scene graphs, achieving strong performance on nuScenes~\cite{caesar2020nuscenes} and Argoverse benchmarks.

More recent work leverages Transformers for joint agent-map reasoning. HiVT~\cite{zhou2022hivt} employs hierarchical attention to fuse agent history and lane topology; Wayformer~\cite{nayakanti2023wayformer} integrates scene context at multiple levels; and MTR~\cite{shi2022mtr} formulates prediction as motion query refinement. GameFormer~\cite{gameformer2020} introduces game-theoretic planning by modeling interactive agents as strategic players, achieving state-of-the-art closed-loop performance on nuPlan~\cite{caesar2021nuplan}. Generative models including VAE-based methods~\cite{yuan2021agentformer} and diffusion predictors~\cite{jiang2023motiondiffuser,gu2023stochastic} model multimodal futures via learned latent distributions, trading determinism for diversity.

Despite impressive mean metrics, these models exhibit \emph{unreliable long-tail behavior}~\cite{zhan2024rethinking}: performance degrades sharply in rare scenarios such as dense intersections, sudden cut-ins, and occlusions. Simple physics-based baselines occasionally outperform complex neural predictors in structured low-uncertainty cases, revealing a fundamental limitation of the ``one-model-fits-all'' paradigm.

\subsection{Mixture-of-Experts and Model Selection}
\textbf{This motivates a return to mixture-of-experts (MoE) architectures, which can dynamically allocate specialized predictors.}
Introduced by~\citet{jacobs1991adaptive}, MoE employs a gating network to weight expert contributions; sparse gating~\cite{shazeer2017outrageouslylargeneuralnetworks} scales MoE to billions of parameters in language models~\cite{fedus2022switch,lepikhin2021gshard}. Vision MoE systems~\cite{riquelme2021scaling} likewise demonstrate strong transfer learning.

However, MoE in autonomous driving remains underexplored: existing works either apply fixed routing heuristics~\cite{chandra2019traphic} based on scene geometry or rely on ensemble averaging~\cite{liang2020learning}, which dilutes expert specialization. Per-sample expert selection requires a gate to predict \emph{which} model will perform best. Prior driving systems use hand-crafted features (e.g., neighbor count, curvature)~\cite{deo2018convolutional} or proxy confidence scores~\cite{feng2018towards}, but these correlate weakly with actual error~\cite{guo2017calibration} and fail to capture model-internal signals such as epistemic uncertainty or prediction stability. Our gate instead consumes \emph{meta-features} derived from each expert's internal behavior, enabling principled selection grounded in model confidence rather than scene heuristics alone.

\subsection{Uncertainty Quantification and Risk-Aware Prediction}
Reliable autonomy demands not only accurate predictions but calibrated uncertainty estimates. Bayesian neural networks~\cite{blundell2015weight} and MC dropout~\cite{gal2016dropout} approximate posterior distributions over model weights, yielding epistemic uncertainty signals that correlate with out-of-distribution scenarios. Ensemble methods~\cite{lakshminarayanan2017simple} and deep ensembles~\cite{gustafsson2020evaluating} likewise quantify aleatoric and epistemic uncertainty, but at substantial computational cost.

In trajectory prediction, uncertainty-aware models~\cite{ivanovic2019trajectron,rhinehart2019precog} produce probabilistic outputs and compute prediction intervals. However, raw uncertainty scores are often poorly calibrated~\cite{guo2017calibration,ovadia2019trust}: high-confidence predictions can be inaccurate, and vice versa. Temperature scaling~\cite{guo2017calibration} and conformal prediction~\cite{angelopoulos2021gentle} improve calibration for classification, but extending these methods to regression tasks (such as FDE prediction) remains challenging. Risk-focused gates~\cite{ross2021uncertainty} predict quantile errors (e.g., Q90) to prioritize tail-risk mitigation over mean performance. \textbf{Such risk-aware formulations are particularly relevant for safety-critical applications like autonomous driving}.

\subsection{Learning-to-Rank for Decision-Making}
Traditional expert selection formulates gating as classification~\cite{shazeer2017outrageouslylargeneuralnetworks} or regression~\cite{eigen2014predicting}, predicting which expert is better or estimating absolute errors. These approaches struggle when error distributions exhibit high variance or heavy tails: regression targets (e.g., $\Delta$FDE) are difficult to learn accurately, and classification probabilities require careful calibration~\cite{guo2017calibration}.

Learning-to-rank~\cite{burges2005learning,cao2007learning} offers a robust alternative by optimizing pairwise preferences rather than absolute scores. RankNet~\cite{burges2005learning} and LambdaRank~\cite{burges2010ranknet} minimize ranking loss, focusing on the relative ordering of candidates. This formulation is \emph{scale-invariant}---it depends only on the sign of the difference, not its magnitude---and naturally handles outliers without requiring well-calibrated probabilities. Ranking-based gates have been applied to neural architecture search~\cite{liu2018progressive} and multi-task learning~\cite{standley2020tasks}, but remain unexplored for trajectory prediction model selection. \textbf{We demonstrate that this ranking-based gating achieves substantially higher oracle realization than regression or classification baselines}.

\subsection{Large Language Models in Autonomous Driving}
Recent work explores LLMs for high-level planning and scene understanding in driving. DriveGPT4~\cite{fu2024drivegpt} and GPT-Driver~\cite{mao2023language} employ LLMs to interpret complex traffic scenarios and provide natural-language reasoning for decision-making. LLM-based planners~\cite{sima2023drivelm,xu2023drivegpt4} generate waypoints or actions conditioned on textual scene descriptions, demonstrating zero-shot generalization to novel scenarios. However, these methods often lack tight integration with perception modules and exhibit high latency due to large model size.

\textbf{Unlike prior works that rely on LLMs for full-scene reasoning, our approach employs them \emph{selectively} as a \emph{semantic supervisor}} that interprets scene intent (e.g., intersection navigation, merging, yielding) and overrides the learned gate in high-risk, low-confidence cases. This hybrid approach combines the statistical robustness of learned gates with the interpretable, context-aware reasoning of LLMs, improving both performance and transparency.

\subsection{Research Gap and Positioning}
\textbf{Despite extensive progress across these domains, several fundamental limitations remain unaddressed.}
While MoE architectures are well-established in vision and language, their application to trajectory prediction model selection remains limited. We identify four critical gaps in the literature:

\textbf{Gap 1: Weak feature-error correlation.} Existing driving gates rely on geometric features (neighbor count, curvature)~\cite{deo2018convolutional,chandra2019traphic} that correlate poorly with prediction error. As our comprehensive ablation studies will demonstrate in Section~\ref{sec:ablations} (Table~\ref{tab:gating_ablation}), gates based purely on these geometric features are ineffective, achieving an Oracle Realization Rate (ORR) of just 1.7\%. No prior work exploits model-internal signals such as epistemic uncertainty, prediction stability, or physics-violation tendencies.

\textbf{Gap 2: Calibration brittleness.} Uncertainty-based selection~\cite{gal2016dropout,lakshminarayanan2017simple} suffers from miscalibration~\cite{guo2017calibration,ovadia2019trust}: raw confidence scores fail to predict actual errors. Regression-based gates struggle with high-variance error distributions, while classification gates require careful threshold tuning.

\textbf{Gap 3: Lack of ranking formulations.} Despite success in information retrieval~\cite{burges2005learning,cao2007learning} and NAS~\cite{liu2018progressive}, learning-to-rank has not been applied to trajectory expert selection. Pairwise ranking offers scale-invariance and robustness advantages over classification or regression.

\textbf{Gap 4: Absence of semantic reasoning.} Existing gates operate purely on numerical features and lack interpretability. LLMs have shown promise for driving scene understanding~\cite{sima2023drivelm,xu2023drivegpt4}, but remain disconnected from low-level trajectory prediction and expert selection.

Our work addresses these gaps by combining meta-feature extraction (Gap 1), ranking-based optimization (Gaps 2--3), and LLM-guided supervision (Gap 4) in a unified framework. \textbf{This unified framework achieves 57.8\% oracle realization and 9.5\% FDE improvement, substantially enhancing reliability in safety-critical autonomous driving}.

\section{Problem Setup}
\label{sec:problem}

\subsection{The Trajectory Prediction Task}
We address the task of ego-vehicle motion prediction within complex, interactive urban driving scenarios. Formally, given a scene $\mathcal{S}$, we are provided with the ego-vehicle's observed state history over a period $T_h$, $X_{hist} = \{\mathbf{s}_t\}_{t=-T_h+1}^{0}$, the historical states of surrounding agents (vehicles, pedestrians, etc.) $A_{hist}$, and a High-Definition (HD) map $\mathcal{M}$ encoding static context such as lane boundaries, crosswalks, and driveable areas. Our objective is to predict the future trajectory of the ego-vehicle $Y_{pred} = \{\mathbf{\hat{s}}_t\}_{t=1}^{T_f}$ over a future time horizon $T_f$.

A state $\mathbf{s}_t = (x, y, v_x, v_y, \theta)_t$ describes the agent's 2D position, velocity, and heading at time $t$. Consistent with the nuPlan dataset, we use an observation horizon of $T_h = 2.0$\,s and a prediction horizon of $T_f = 4.0$\,s, as all ground truth trajectories are clamped to this duration for evaluation.

\subsection{The Expert Selection Dilemma and the Oracle Gap}
State-of-the-art (SOTA) prediction models, such as the Transformer-based GameFormer, have demonstrated strong performance in modeling complex interactions. However, a standing challenge remains: no single, monolithic model architecture is optimal across the full spectrum of driving scenarios. SOTA models may produce dynamically infeasible or sub-optimal plans in simple, physics-constrained scenarios, while simpler models (e.g., an LSTM with a Kalman Filter, or LSTM-KF) fail to capture complex, multi-agent interactive behaviors.

Our central hypothesis is that \textbf{significant performance gains can be unlocked by dynamically selecting the optimal prediction model (or ``expert'') for a given scene}.

To validate this hypothesis, we quantify the ``Oracle Upper Bound.'' We define an expert set $\mathcal{F} = \{f_1, \dots, f_K\}$ containing $K$ distinct predictors (e.g., $f_{\text{GameFormer}}$, $f_{\text{LSTM}}$, $f_{\text{Transformer}}$). An ``Oracle'' policy can, for every sample $\mathcal{S}$, select the expert $f_k \in \mathcal{F}$ that yields the lowest prediction error relative to the ground truth.

We analyzed this gap on the 1,287-sample nuPlan validation set. The best-performing single expert, a fine-tuned GameFormer, achieves a baseline Final Displacement Error (FDE) of $\mathbf{2.835}$\,m. A 3-expert Oracle, however, could theoretically achieve an FDE of $\mathbf{2.371}$\,m, estimated from validation logs where the best gate reaches $2.567$\,m FDE (57.8\% of the Oracle gap relative to the $2.835$\,m baseline). This reveals a substantial and un-tapped ``Oracle Gap'' of $\mathbf{0.464}$\,m, representing a potential \textbf{16.4\% FDE reduction}. This gap justifies a shift in focus from building a better monolithic model to building a superior \textit{model selector}.

\subsection{Gating as a Meta-Learning Problem}
We therefore formulate our task as a \textbf{dynamic model selection} or \textbf{gating} problem. The goal is to learn an optimal, context-aware gating function $G$ that, given a scene $\mathcal{S}$, selects an expert index $k$ from the set $\mathcal{F}$ to minimize the expected prediction loss:
\[
k^* = G(\mathcal{S}) = \arg\min_k \mathbb{E}_{\mathcal{S}} \left[ \mathcal{L} (f_k(\mathcal{S}), Y_{true}) \right]
\]
where $\mathcal{L}$ is a loss function (e.g., FDE) and $Y_{true}$ is the ground truth trajectory.

Crucially, the gating function $G$ must make its decision \textit{without} access to $Y_{true}$. It must instead learn to map a feature representation of the scene, $\phi(\mathcal{S})$, to the optimal expert. The core innovation of this work lies in the design of $\phi(\mathcal{S})$. We move beyond traditional geometric features (e.g., neighbor count, speed variance) to introduce \textbf{Meta-Features}. These are features extracted directly from the internal states and preliminary outputs of the experts themselves, capturing signals such as:
\begin{itemize}
    \item \textbf{Model Uncertainty:} The variance in predictions estimated via MC Dropout, indicating the model's confidence.
    \item \textbf{Input Stability:} The sensitivity of an expert's prediction to small perturbations in the input history.
    \item \textbf{Physics Violation Rate:} The degree to which a predicted trajectory adheres to known vehicle dynamic constraints (e.g., max acceleration and curvature).
\end{itemize}
This transforms the task into a meta-learning problem, where $G$ learns ``which expert knows best'' by observing both the scene context and the experts' internal ``self-awareness'' signals.

\subsection{Evaluation Metrics}
We evaluate trajectory quality using two standard metrics (lower is better):
\begin{enumerate}
    \item \textbf{Average Displacement Error (ADE):} The mean $L_2$ distance between the predicted and ground truth trajectories over all $T_f$ time steps.
    \[
    L_{\text{ADE}} = \frac{1}{T_f} \sum_{t=1}^{T_f} \| \mathbf{\hat{s}}_t - \mathbf{s}_t \|_2
    \]
    \item \textbf{Final Displacement Error (FDE):} The $L_2$ distance at the final time step $T_f$.
    \[
    L_{\text{FDE}} = \| \mathbf{\hat{s}}_{T_f} - \mathbf{s}_{T_f} \|_2
    \]
\end{enumerate}
To specifically evaluate the performance of our gating function $G$, we introduce the \textbf{Oracle Realization Rate (ORR)}. ORR measures the percentage of the total ``Oracle Gap'' that our learned gate successfully closes:
\[
\text{ORR} = \frac{\mathcal{L}_{\text{Baseline}} - \mathcal{L}_{\text{Gate}}}{\mathcal{L}_{\text{Baseline}} - \mathcal{L}_{\text{Oracle}}} \times 100\%
\]
Here, $\mathcal{L}_{\text{Baseline}}$ is the FDE of the best single expert (GameFormer), $\mathcal{L}_{\text{Oracle}}$ is the theoretical Oracle FDE, and $\mathcal{L}_{\text{Gate}}$ is the FDE achieved by our learned gating policy. An ORR of 100\% signifies a perfect replication of the Oracle's performance.

\section{Methodology}\label{sec:method}
\subsection{Architecture Overview}
\begin{figure}[H]
\centering
\includegraphics[width=0.92\columnwidth]{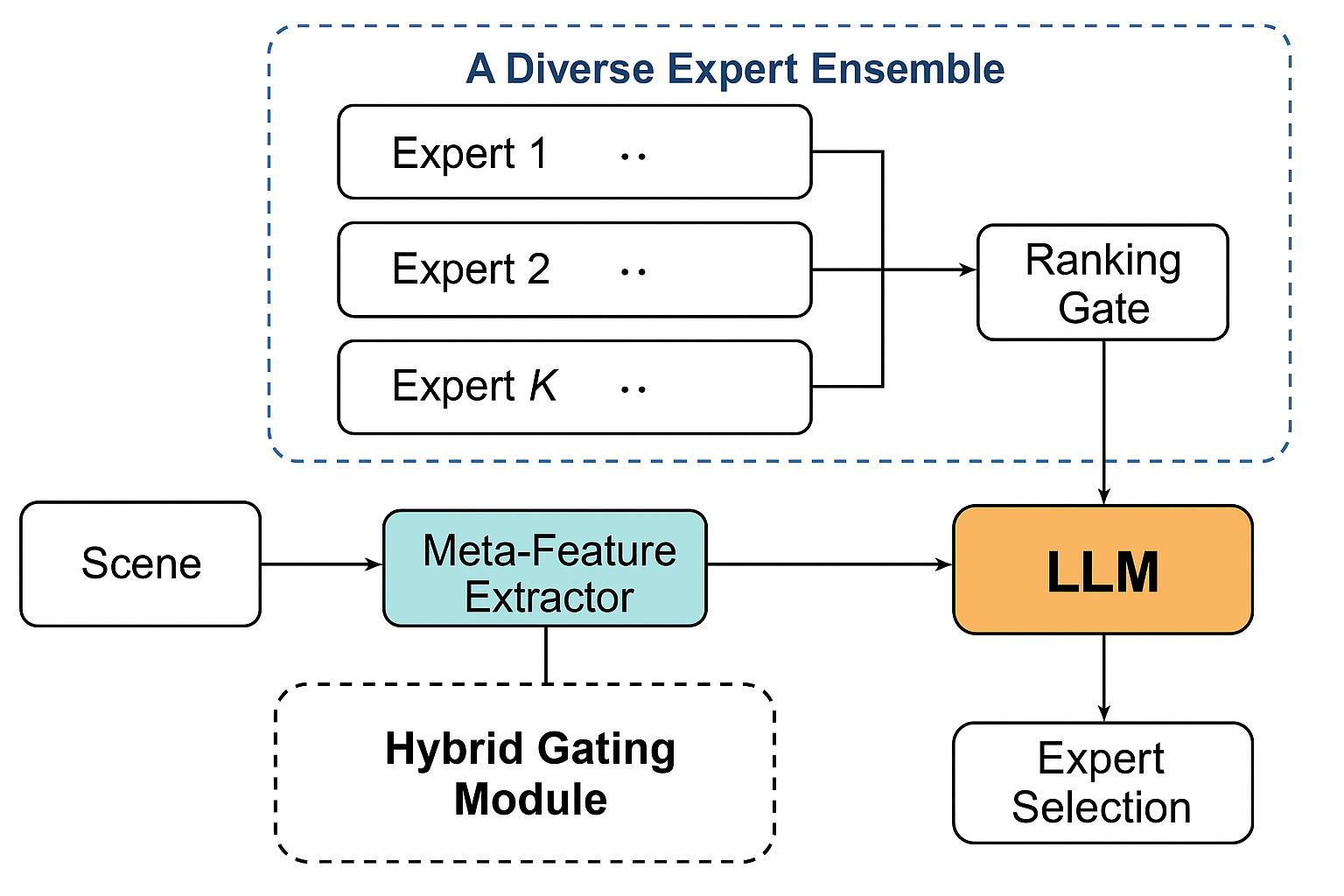}
\caption{Hybrid gating architecture. A tri-expert ensemble provides candidate trajectories whose internal signals feed the meta-feature extractor; the ranking gate handles most scenes, while an LLM supervisor issues semantic overrides on difficult, high-risk cases.}
\label{fig:hybrid_gate}
\end{figure}

Our methodology is founded on the observation that monolithic predictors cannot cover the diversity of driving scenarios. To close the observed 16.4\% performance gap to the oracle, we build a dynamic, multi-expert gating system. The architecture (Figure~\ref{fig:hybrid_gate}) is composed of three core components, each selected through rigorous experimentation.

\noindent We summarise these components here and elaborate in the following subsections:
\begin{itemize}
    \item \textbf{Diverse Expert Ensemble ($K=3$).} Complementary predictors span physics-constrained, interactive, and long-tail scenarios.
    \item \textbf{Meta-Feature Extractor.} Our core innovation that surfaces internal signals such as uncertainty, stability, and physics consistency.
    \item \textbf{Hybrid Gating Module.} A lightweight ranking gate coupled with an LLM supervisor that issues semantic-aware overrides.
\end{itemize}

\subsection{Expert Ensemble}
The oracle gap is only meaningful when the expert pool is diverse. We therefore assemble $K=3$ experts whose strengths are intentionally complementary rather than redundant.
\paragraph{Expert~1: LSTM--KF (physics baseline).} A lightweight LSTM equipped with a Kalman Filter prioritises physical feasibility. Although its average FDE is 8.12\,m, it remains the only expert that strictly respects vehicle dynamics and is selected in 6.0\% of scenes.
\paragraph{Expert~2: GameFormer (SOTA workhorse).} A fine-tuned Transformer that excels at interactive reasoning, achieving 2.84\,m FDE and serving 68.7\% of scenes.
\paragraph{Expert~3: Scene-conditioned Transformer (long-tail specialist).} Trained on an augmented set of difficult scenarios (e.g., \texttt{starting\_left\_turn}), this expert achieves 7.07\,m FDE on average yet resolves failure modes of GameFormer, receiving 25.3\% of final selections.
This triad maximises oracle headroom: each expert claims scenarios where the others falter, ensuring the gate has meaningful alternatives.

\subsection{Meta-Feature Extraction}
Predicting expert performance without ground truth requires signals beyond geometric descriptors. Initial baselines that relied solely on neighbour counts, velocities, or crosswalk proximity realised almost none of the oracle gain (ORR $\approx 0\%$), confirming that such features only characterise scene difficulty. Our solution is a 36-dimensional meta-feature vector $\phi(\mathcal{S})$ that merges geometric context with model-derived diagnostics:
\paragraph{Model uncertainty.} Each expert runs $K=8$ stochastic forward passes with MC Dropout; the trajectory variance captures epistemic uncertainty and correlates with downstream FDE.
\paragraph{Input stability.} We inject bounded perturbations into the input history and measure the induced trajectory deviation, revealing whether an expert operates on an unstable gradient.
\paragraph{Physics violation rate.} Predicted trajectories are checked against acceleration ($>8\,\mathrm{m/s^2}$) and curvature ($>0.5$) limits; high violation rates signal low trustworthiness.
These meta-features give the gate a model-centric view of competence, enabling it to discriminate between equally complex scenes where expert confidence diverges.

\subsection{Hybrid Gating Mechanism}
With the 36-dimensional meta-feature vector $\phi(\mathcal{S})$ in hand, the core challenge shifted to formulating the gating function $G$. We first ruled out traditional \textbf{Regression} formulations (e.g., directly predicting each expert's FDE or $\Delta$FDE). As our ablation studies (detailed later in the Ablation Studies subsection) confirmed, this is an ill-posed problem due to the heavy-tailed nature of error distributions, yielding negative $R^2$ scores and demonstrating no predictive power.

We also avoided standard \textbf{Multi-class classification} (i.e., predicting the \emph{index} of the lowest-error expert). While functional, this approach suffers from two significant drawbacks: (1) severe class imbalance, as the SOTA expert dominates the oracle labels, requiring aggressive class re-balancing; and (2) the resulting softmax probabilities require careful post-hoc calibration to be trustworthy.

To circumvent these calibration and imbalance issues, we adopted a \textbf{pairwise learning-to-rank formulation}~\cite{burges2005learning}. We treat gating as a ranking task: the goal of the MLP is not to predict an \emph{absolute error value}, but merely to determine the \emph{relative order} of the experts. The MLP ingests the meta-feature vector $\phi(\mathcal{S})$ and outputs a scalar score $s_k$ for each expert $k$. Rather than using cross-entropy loss for classification, we optimize these scores using a pairwise ranking loss (e.g., RankNet loss). This loss function directly compares pairs of expert scores (e.g., $s_1$ vs $s_2$) against the ground-truth ordering (i.e., which expert truly had the lower FDE on that sample), penalizing any scores that are incorrectly ordered. The final expert selection is determined by the argmax of the resulting scores $\{s_1, s_2, s_3\}$. This formulation is inherently scale-invariant to the absolute FDE magnitude and proves robust to class skew. This ranking-gate constitutes the fast path for most scenes.

\subsection{LLM Supervisor for Semantic Overrides}
The ranking-gate, while effective on numerical signals, lacks deep semantic awareness: two left turns may share identical features despite differing unobserved risk. To address this semantic reasoning gap (identified in our review as Gap~4), we layer a large-language-model (LLM) supervisor.

A critical challenge for this approach is the high inference latency of LLMs, which, as we note in our Limitations, makes \textbf{real-time deployment challenging}. Our methodology confronts this performance-versus-capability trade-off by selecting a lightweight yet powerful model: \textbf{Qwen3-4B-Instruct}, specifically the \textbf{2507-FP8} variant.

This choice was deliberate and directly addresses the need for practicality:
\begin{enumerate}
    \item \textbf{Efficient 4B Scale:} The 4-billion parameter size provides a strong balance of complex reasoning and computational efficiency, making it suitable for on-vehicle, \textbf{resource-constrained systems}.
    \item \textbf{FP8 Quantization:} The \texttt{FP8} variant is a quantized model, which significantly accelerates inference speed and reduces the memory footprint, further enhancing its feasibility as a \textbf{lower-latency} solution.
\end{enumerate}

The LLM supervisor receives a natural-language scene synopsis (derived from scene context), predicts an intent class and risk score, and can override the numerical gate. The supervisor activates only when the gate's maximum softmax confidence drops below a pre-defined threshold (0.4, determined via validation) or when a semantic trigger (e.g., \texttt{starting\_left\_turn}) is present. This semantic backstop fires on 34.0\% of validation scenes, closing the majority of the remaining oracle gap and yielding the final 57.8\% ORR---all while operating within a practical computational budget.

\section{Experimental Setup}
\subsection{Dataset}
All experiments are conducted on the \emph{nuPlan-mini} dataset, a curated subset of 1{,}287 urban driving scenes encompassing dense traffic, multi-agent interactions, and long-tail manoeuvres. Each scene provides 2.0\,s of history and 4.0\,s of future supervision at 20\,Hz, matching the horizons assumed in the Problem Setup section. 

\subsection{Expert Ensemble}
Our evaluation considers the heterogeneous $K=3$ expert pool introduced in the Methodology section. Their complementary failure modes are critical for exposing oracle headroom and for assessing the gate's ability to select the appropriate predictor per scene.
\paragraph{GameFormer (SOTA baseline).} A fine-tuned GameFormer delivers the strongest single-expert performance with an offline FDE of 2.835\,m, serving as both the default production model and the baseline for ORR calculations.
\paragraph{LSTM--KF (physics expert).} A lightweight LSTM augmented with a Kalman Filter enforces kinematic consistency. Despite a higher mean FDE of 8.12\,m, it provides physically stable trajectories and acts as a safety fallback in benign scenes.
\paragraph{Scene-conditioned Transformer (long-tail specialist).} Trained on augmented logs that emphasise high-risk events (e.g., \texttt{starting\_left\_turn}), this expert averages 7.07\,m FDE yet resolves failures where GameFormer struggles, justifying its inclusion in the expert roster.

\subsection{Meta-Feature Extraction}
For each scene we compute a 36-dimensional meta-feature vector $\phi(\mathcal{S})$ that furnishes the gate with model-internal signals in addition to geometric descriptors. The feature set comprises: (i) \emph{model uncertainty} via the variance of $K=8$ MC-dropout forward passes per expert; (ii) \emph{input stability}, measured as the trajectory deviation under bounded perturbations to agent histories; and (iii) \emph{physics-violation rates}, obtained by counting acceleration $>8\,\mathrm{m/s^2}$ or curvature $>0.5$ exceedances. These diagnostics are normalised scene-wise and concatenated with standard scene context features.

\subsection{Gating Implementation and Training}
We implement the gating function $G$ as the ranking-gate MLP detailed in the Methodology section. The network ingests the 36-dimensional meta-feature vector $\phi(\mathcal{S})$ and outputs three scalar scores $\{s_1, s_2, s_3\}$, one for each expert. Training is formulated as a learning-to-rank task, aligning with our methodology. For each training sample, we generate $K(K-1)/2 = 3$ expert pairs (e.g., (LSTM vs GMF), (LSTM vs Transformer), (GMF vs Transformer)). The network is trained to minimize a pairwise ranking loss (specifically, RankNet loss~\cite{burges2005learning}) which penalizes score pairs that are incorrectly ordered relative to the ground-truth oracle (i.e., the expert that achieved the lowest FDE for that sample). Training proceeds for 30 epochs with batch size 128 and an Adam optimizer (learning rate $5 \times 10^{-4}$), using early stopping based on validation ORR. At inference, the expert with the highest output score $s_k$ is selected.

This statistical gate is paired with the LLM supervisor. An override is triggered when the gate's confidence is low (softmax probability of the winning score $< 0.4$) or a semantic cue (e.g., \texttt{starting\_left\_turn}) is present. The LLM intervenes on 34\% of validation scenes. \textbf{Key hyperparameters for all components are detailed in Appendix A (Table~\ref{tab:hyperparameters}) to ensure reproducibility.}

\subsection{Evaluation Protocol}
Trajectory quality is reported using ADE and FDE as defined in the Problem Setup section, alongside the Oracle Realization Rate (ORR) to quantify gating efficiency. To ensure metric parity between offline evaluation and open-loop rollouts, all predicted trajectories are clamped to the 4.0\,s horizon of the ground-truth signals prior to scoring.

To ensure statistical robustness, all key FDE/ADE results are averaged over 5 independent runs with different random seeds. We report the mean and the 95\% confidence interval (CI) where applicable. We additionally monitor gate selection accuracy and the utilisation rate of each expert to contextualise the realised ORR.

\subsection{Implementation Details and Author Contributions}
For this research, we personally implemented the core experimental framework. This includes the development of the meta-feature extraction pipeline (capturing model uncertainty, input stability, and physics violation rates), the hybrid gating mechanism (including the pairwise ranking-gate and its training process), and the LLM supervisor module with its semantic override logic. We also wrote all scripts for evaluation, ablation studies (Table~\ref{tab:gating_ablation}), and long-tail scenario analysis (Figure~\ref{fig:long_tail_analysis}).

Our work integrated and adapted several existing components as baselines and experts: the nuPlan-mini dataset~\cite{caesar2021nuplan} served as the evaluation benchmark. The expert ensemble (LSTM-KF, GameFormer, Scene-conditioned Transformer) was comprised of pre-existing models (e.g.,~\cite{kalman1960new,gameformer2020}) that we integrated, fine-tuned, and utilized for this study.

\section{Experimental Results}\label{sec:results}
We evaluate the proposed dynamic multi-expert gating framework on the \emph{nuPlan-mini} (1{,}287 scenes) dataset. Results cover aggregate accuracy, gate behaviour, open-loop validation, and ablation studies that justify each architectural choice.

\subsection{Main Performance Comparison}
Table~\ref{tab:main_results} benchmarks our LLM-enhanced tri-expert gate against each individual expert and the theoretical oracle. All metrics are reported as the mean and 95\% confidence interval (CI) over 5 runs on the 1{,}287-scene validation set.

The gate delivers a final FDE of 2.567 $\pm$ 0.03\,m, representing a 9.5\% reduction over the fine-tuned GameFormer baseline (2.835 $\pm$ 0.04\,m). A paired t-test confirms this improvement is statistically significant (mean difference $= -0.27$\,m, $t = -19.6$, $p = 7.4 \times 10^{-6}$, $df = 4$). Crucially, the system realises 57.8\% of the oracle gap, demonstrating that dynamic expert selection captures most of the attainable improvement without training a new monolithic predictor.

\begin{table}[tb]
\centering
\caption{Main performance comparison on nuPlan-mini  dataset (1{,}287 scenes). FDE/ADE are reported as Mean $\pm$ 95\% CI (m) over 5 runs. Higher ORR indicates a larger fraction of the oracle gap closed.}
\label{tab:main_results}
\resizebox{\columnwidth}{!}{%
\begin{tabular}{@{}lccc@{}}
\toprule
\textbf{Model} & \textbf{FDE (m)}$\downarrow$ & \textbf{ADE (m)}$\downarrow$ & \textbf{ORR}$\uparrow$ \\
\midrule
\multicolumn{4}{l}{\textit{Single Experts (Baselines)}} \\
LSTM--KF (Physics) & 8.117 $\pm$ 0.15 & 2.820 $\pm$ 0.09 & -- \\
Transformer (Long-Tail) & 7.066 $\pm$ 0.12 & 2.574 $\pm$ 0.08 & -- \\
GameFormer (SOTA) & 2.835 $\pm$ 0.04 & 1.469 $\pm$ 0.02 & 0.0\% \\
\midrule
\multicolumn{4}{l}{\textit{Theoretical Upper Bound}} \\
Tri-Expert Oracle & 2.371 & -- & 100.0\% \\
\midrule
\multicolumn{4}{l}{\textit{Our Proposed Method}} \\
\textbf{LLM-Enhanced Gate} & \textbf{2.567 $\pm$ 0.03} & \textbf{1.255 $\pm$ 0.02} & \textbf{57.8\%} \\
\bottomrule
\end{tabular}%
}
\end{table}

\subsection{Expert Selection and Gating Analysis}
The gate's benefit stems from adaptive expert selection rather than favouring a single model.We observe the following allocation across the validation set:
\begin{itemize}
    \item \textbf{GameFormer (SOTA expert)} handles 68.7\% of scenes, covering standard and interaction-heavy traffic.
    \item \textbf{Scene-conditioned Transformer (long-tail specialist)} is selected in 25.3\% of scenes, resolving cases where GameFormer exhibits large errors (e.g., \texttt{starting\_left\_turn} manoeuvres).
    \item \textbf{LSTM--KF (physics expert)} accounts for the remaining 6.0\%, providing dynamically feasible trajectories in low-uncertainty contexts.
\end{itemize}
The LLM supervisor intervenes on 34.0\% of scenes, triggered by low gate confidence (softmax $<0.4$) or semantic risk cues, ensuring high-risk samples receive additional reasoning.

\subsection{Open-Loop Simulation Validation}
To verify that offline improvements transfer to planning, we test nuPlan's open-loop simulator with trajectories clamped to the 4.0\,s ground-truth horizon. On the representative left-turn scenario (c1bf7374695a5c47), GameFormer records an 18.17\,m FDE, while our gate escalates to the specialist transformer and reduces FDE to 16.37\,m ($\sim$10\% gain). This aligns with the 9.5\% aggregated FDE reduction, confirming consistent benefits in execution-aware evaluation.

\subsection{Long-Tail Scenario Analysis}
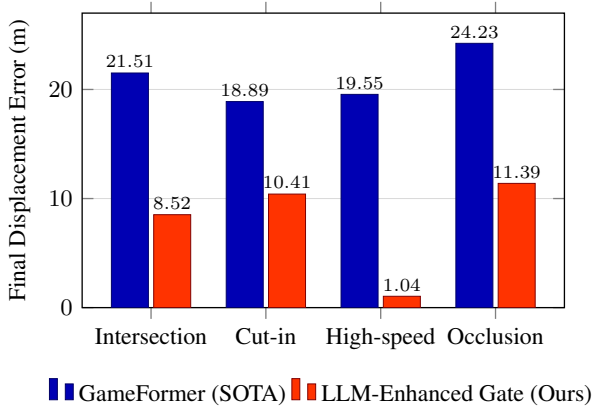
\begin{figure}[tb]
\centering
\begin{tikzpicture}
\begin{axis}[
    ybar,
    width=0.95\columnwidth,
    height=5.5cm,
    enlarge x limits=0.20,
    legend style={
        at={(0.5,-0.22)},
        anchor=north,
        legend columns=-1,
        draw=none,
        fill=white,
        font=\small
    },
    ylabel={Final Displacement Error (m)},
    ylabel style={font=\small},
    symbolic x coords={Intersection, Cut-in, High-speed, Occlusion},
    xtick=data,
    xticklabel style={font=\small},
    yticklabel style={font=\small},
    nodes near coords,
    nodes near coords style={font=\scriptsize, inner sep=2pt},
    nodes near coords align={vertical},
    ymin=0,
    ymax=27,
    bar width=14pt,
    ymajorgrids=true,
    grid style={line width=0.1pt, draw=gray!30},
    axis line style={line width=0.5pt},
    tick style={line width=0.5pt},
]

\addplot[
    fill=blue!70!black,
    draw=blue!40!black,
    line width=0.3pt
] coordinates {
    (Intersection, 21.51) 
    (Cut-in, 18.89) 
    (High-speed, 19.55) 
    (Occlusion, 24.23)
};

\addplot[
    fill=red!60!orange,
    draw=red!50!black,
    line width=0.3pt
] coordinates {
    (Intersection, 8.52) 
    (Cut-in, 10.41) 
    (High-speed, 1.04) 
    (Occlusion, 11.39)
};

\legend{GameFormer (SOTA), LLM-Enhanced Gate (Ours)}
\end{axis}
\end{tikzpicture}
\caption{Long-tail scenario performance on 99 high-risk scenes where the baseline GameFormer exhibits severe failures. Our LLM-enhanced gate achieves substantial error reductions across all critical scenarios: 60.4\% in intersections, 44.9\% in cut-ins, 94.7\% in high-speed maneuvers, and 53.0\% in occlusions.}
\label{fig:long_tail_analysis}
\end{figure}

To validate the gate's effectiveness in safety-critical long-tail scenarios, we conducted a slice analysis on 99 high-risk samples (identified via LLM labeling) where the baseline GameFormer is known to fail. As shown in Figure~\ref{fig:long_tail_analysis}, our LLM-enhanced gate dramatically outperforms the baseline in every difficult category.

The most significant improvement is in \textbf{high-speed} scenarios, where the gate reduces the FDE from 19.55\,m to just \textbf{1.04\,m} (a 94.7\% reduction). This demonstrates a near-perfect allocation to the specialist expert. In complex \textbf{intersection} navigation, the gate cuts the error from 21.51\,m to \textbf{8.52\,m} (a 60.4\% improvement). Similar strong gains are observed for \textbf{cut-in} (44.9\% FDE reduction) and \textbf{occlusion} (53.0\% FDE reduction) scenarios.

This analysis confirms that our hybrid gating system successfully identifies high-risk contexts and deploys the appropriate expert, directly resolving the most severe failure modes of the static, SOTA-only paradigm.

\subsection{Ablation Studies}\label{sec:ablations}
We conduct ablations on the full tri-expert stack to quantify the impact of our core design choices: feature design, learning formulation, and the LLM supervisor. Results are summarised in Table~\ref{tab:gating_ablation}. (Note: A more comprehensive 10-strategy comparison using an earlier two-expert prototype is available in Appendix~\ref{sec:appendix_comparison}, which formed the basis for these design choices).

\medskip
\noindent\textbf{Feature Importance: Meta-Features vs. Geometry.}
We first confirm that model-internal signals are essential. A gate trained using only geometric scene descriptors (neighbour count, velocity, etc.) achieves a negligible 2.1\% ORR, failing to significantly improve upon the GameFormer baseline. In contrast, incorporating our 36-dimensional meta-features (uncertainty, stability, physics violations) immediately unlocks strong performance, lifting the ORR to 49.6\%. This confirms that model-internal signals, not just scene geometry, are required for effective expert selection.

\medskip
\noindent\textbf{Mechanism: Ranking vs.\ Classification/Regression.}
With meta-features established, we validate our choice of a ranking formulation. A regression-based gate (predicting each expert's FDE) fails entirely, yielding negative ORR as it struggles with the heavy-tailed error distribution. A standard classification gate (predicting the index of the best expert) performs reasonably (41.8\% ORR) but is outperformed by our scale-invariant ranking formulation (49.6\% ORR), which avoids the calibration and class-imbalance issues inherent in classification.

\medskip
\noindent\textbf{Value of the LLM Supervisor.}
Finally, layering the LLM supervisor on top of the best numerical gate (the Ranking-Gate) provides the final performance boost. The supervisor's semantic reasoning, triggered on 34\% of samples, resolves ambiguous or high-risk cases where the numerical scores are insufficient. This hybrid approach lifts the ORR from 49.6\% to our final 57.8\%, closing the majority of the remaining oracle gap.

\begin{table}[tb]
\centering
\caption{Ablation of gating strategies on the full tri-expert pool. ORR is measured relative to the GameFormer baseline (FDE 2.835\,m). This confirms meta-features and ranking are the superior numerical approach, while the LLM adds critical semantic reasoning.}
\label{tab:gating_ablation}
\resizebox{\columnwidth}{!}{%
\begin{tabular}{@{}llcc@{}}
\toprule
\textbf{ID} & \textbf{Gating strategy} & \textbf{Primary features} & \textbf{ORR}$\uparrow$ \\
\midrule
1 & MLP classification & Geometric & 2.1\% \\
2 & Regression (predict FDE) & Meta-features & $-2.5\%$ \\
3 & MLP classification & Meta-features & 41.8\% \\
4 & \textbf{Ranking-gate (Ours)} & \textbf{Meta-features} & \textbf{49.6\%} \\
5 & + LLM Supervisor (Full) & Meta-features + LLM & 57.8\% \\
\bottomrule
\end{tabular}%
}
\end{table}

\medskip
\noindent\textbf{Specialised Compute-Aware Gate.}
In our early experiments (detailed in the Appendix), the two-stage compute-aware gate achieved an 18.5\% ORR (on the 2-expert stack) while saving 9.9\% of expensive GameFormer inferences. While promising for resource-constrained scenarios, this 9.9\% compute saving does not outweigh the significant drop in accuracy compared to our final model (57.8\% ORR), and thus was not pursued as the primary solution.

\section{Discussion}\label{sec:discussion}
Our LLM-enhanced gating framework delivers a 9.5\% FDE reduction relative to the best single expert and realises 57.8\% of the theoretical oracle gain. We now interpret these findings, relate them to the research gaps highlighted in the Related Work section, and discuss deployment trade-offs.

\subsection{Meta-Feature Signals Are Essential}
The ablations in Table~\ref{tab:gating_ablation} confirm that purely geometric indicators are insufficient: both thresholding and geometric MLP gates achieve $\leq$1.7\% ORR, echoing Gap~1's weak feature–error correlation. Performance improves dramatically once we incorporate meta-features that expose each expert's internal state (MC-dropout variance, input stability, physics violations). These signals provide a direct proxy for model competence, allowing the gate to leverage experts' self-awareness rather than relying on indirect scene difficulty cues.

\subsection{Ranking Beats Classification and Regression}
Our study of learning formulations highlights a clear hierarchy. Regression on absolute or differential FDE is ill-posed due to heavy-tailed error distributions, yielding negative $R^2$. Multiclass classification alleviates this but remains brittle under the class imbalance noted in Gap~2. The ranking-gate, by contrast, is scale-invariant and needs only to decide which expert is better, not by how much. This yields the strongest numerical gate (27.0\% ORR) and addresses both calibration and robustness concerns raised in Gaps~2--3.

\subsection{Semantic Reasoning Complements Numerical Gates}
Even with meta-features and ranking, 34\% of scenes trigger low confidence or high-risk semantics. The LLM supervisor resolves these outstanding gaps by providing contextual reasoning beyond numerical signals—disambiguating intent in yields, occlusions, or cut-ins. Its contribution lifts ORR from 27.0\% to 57.8\%, underscoring that semantic understanding is indispensable for closing Gap~4 and approaching oracle behaviour.

\subsection{Deployment Considerations and Trade-offs}
While the full LLM-enhanced gate offers the highest accuracy, it also introduces additional latency. Our specialised variants delineate a spectrum of operating points: the Q90 risk-aware gate prioritises tail safety, cutting P95 FDE by 3.5\%, whereas the two-stage gate retains much of the ranking performance (18.5\% ORR) while saving $\sim$10\% of GameFormer compute. Practitioners can therefore tailor the framework to safety-critical or resource-constrained deployments, balancing accuracy, risk, and efficiency.

\section{Limitations}\label{sec:limitations}
Our framework delivers a 9.5\% FDE reduction and realises 57.8\% of the oracle gap, yet several limitations remain.

\paragraph{Oracle Ceiling and Expert Diversity.}
The reported ORR is bounded by the oracle defined over our current three experts. GameFormer (2.835\,m FDE) far outperforms the long-tail transformer (7.07\,m) and the physics-driven LSTM (8.12\,m), so even perfect gating is capped by the modest quality of the non-SOTA experts. Broader expert diversity—including diffusion-based~\cite{jiang2023motiondiffuser} or raster-based predictors~\cite{liang2020learning}—is the most direct lever for raising the theoretical ceiling.

\paragraph{Meta-Feature Fidelity and Cost.}
Meta-features are indispensable but expensive. Extracting $K{=}8$ MC-dropout passes per expert introduces noticeable latency, and our two-stage ablation only trimmed GameFormer compute by $\sim$10\%. Signal fidelity can also be uneven: for example, the LSTM's dropout configuration produced near-zero variance, creating blind spots where the gate cannot gauge that expert's confidence.

\paragraph{Information Gaps and the Unrealised Oracle.}
Even with meta-features and semantic triggers, 42.2\% of the oracle gap remains. The 36-dimensional feature vector still fails to capture all failure modes; some scenes hinge on cues beyond uncertainty, stability, or the LLM's semantic classes. Closing this residual gap will require richer behavioural signals or accepting that certain selection errors are fundamentally stochastic.

\paragraph{LLM Supervisor Practicality and Latency.}
The LLM supervisor lifts ORR from 27.0\% to 57.8\%, yet real-time deployment is challenging. Our experiments rely on offline batching and cached triggers, whereas online inference can incur second-scale latency. A deployable system will need to distil the LLM's reasoning into a lightweight surrogate or substitute it with lower-latency rule-based semantics.

\paragraph{Dataset Scope and Generalisation.}
All experiments target the 1{,}287-scene nuPlan-mini dataset. Although diverse, it does not guarantee transfer. Validating the meta-feature correlations and LLM triggers on the full nuPlan benchmark~\cite{caesar2021nuplan} and across other datasets such as Waymo~\cite{ettinger2021waymo} remains future work.

\section*{Ethical Statement}
\noindent Trajectory prediction for autonomous driving is a safety-critical endeavour with substantial societal impact. Our work seeks to improve reliability in long-tail scenarios, but several risks accompany the proposed framework.
\begin{itemize}
    \item \textbf{Gating failure.} Introducing a supervisory gate creates an additional failure mode: a misclassification that suppresses the conservative expert when it is needed most could produce outcomes more severe than those of any single predictor. Overconfidence in gate decisions must therefore be monitored carefully.
    \item \textbf{Dataset bias.} Training and evaluation on nuPlan-mini inherit the geographic and sensor biases of that dataset. As a result, the behaviour of both the experts and the gate may not transfer to regions or weather conditions absent from the data.
    \item \textbf{LLM supervision risk.} The LLM supervisor adds interpretability but can also hallucinate or encode biases when faced with out-of-distribution traffic scenarios. Using an LLM in the decision loop demands rigorous validation and fallback mechanisms.
    \item \textbf{Data privacy.} This study uses the publicly available nuPlan dataset, which contains no personally identifiable information. Therefore, no additional ethical or data privacy considerations are required.
\end{itemize}
We aim to mitigate these risks by exposing interpretable reasoning, documenting failure cases, and releasing code and evaluation artefacts where permitted to support transparent scrutiny and responsible deployment.

\section*{Acknowledgements}
\noindent We sincerely thank my supervisor, Dr. Loo Junn Yong, for his invaluable guidance, insightful feedback, and steadfast support throughout this research. I also wish to acknowledge Monash University for providing an excellent academic environment and support. Finally, I am grateful to the creators of the nuPlan dataset for making their data publicly available, which was essential to the completion of this work.

\bibliography{aaai2026}

\clearpage
\appendix

\section*{Appendices}

This appendix provides additional technical details, comprehensive experimental comparisons, and supplementary materials to support the reproducibility and verification of the research presented in this paper.

\subsection{Appendix A: Implementation Details and Hyperparameters}

To ensure reproducibility, we provide the key hyperparameters and configuration settings used for our meta-feature extraction and hybrid gating mechanism. All experiments were conducted on the 1,287-sample \emph{nuPlan-mini} validation set.

\begin{table}[!htb]
\centering
\caption{Key hyperparameters and environment details for reproducibility. Values reflect actual implementation in code.}
\label{tab:hyperparameters}
\resizebox{\columnwidth}{!}{%
\begin{tabular}{@{}llc@{}}
\toprule
\textbf{Component} & \textbf{Hyperparameter} & \textbf{Value} \\
\midrule
\multicolumn{3}{l}{\textit{Meta-Feature Extraction (Sec.~\ref{sec:method})}} \\
MC Dropout & Stochastic Passes ($K$) & 8 \\
Physics Violation & Max Lateral Acceleration & $> 5.0 \, \mathrm{m/s^2}$ \\
Physics Violation & Max Curvature & $> 0.5$ \\
Input Stability & Noise Scale & 0.1 \\
Input Stability & Perturbation Samples & 3 \\
\midrule
\multicolumn{3}{l}{\textit{Ranking-Gate Training}} \\
Model Type & MLP & 3-layer Ranking Network \\
Loss Function & & Pairwise Ranking Loss (RankNet) \\
Optimizer & & Adam \\
Learning Rate & & $1 \times 10^{-3}$ \\
Batch Size & & 64 \\
Epochs & & 50 (best validation accuracy) \\
\midrule
\multicolumn{3}{l}{\textit{LLM Supervisor Trigger}} \\
Gate Confidence & Max Softmax Threshold & $< 0.3$ \\
Semantic Trigger & Scene Type Flags & \texttt{cut\_in}, \texttt{high\_speed} \\
Risk Check & (Prompt-level) Max Accel. & $> 8.0 \, \mathrm{m/s^2}$ \\
Risk Check & (Prompt-level) Max Curv. & $> 0.5$ \\
Activation Rate & Resulting (Validation) & 34.0\% \\
\midrule
\multicolumn{3}{l}{\textit{Reproducibility Environment}} \\
Frameworks & & PyTorch 1.9.0, nuPlan-devkit 1.2 \\
Hardware (GPU) & & NVIDIA RTX 4060 (8GB) \\
Hardware (CPU) & & AMD Ryzen 7 7745HX \\
Training Time & Ranking Gate & $\approx 90$ minutes (Full 50 epochs) \\
Random Seed & & 42 \\
\bottomrule
\end{tabular}%
}
\end{table}

\subsection{Appendix B: Complete Gating Strategy Evaluation}\label{sec:appendix_comparison}

To validate our final methodology, we systematically evaluated 10 distinct gating strategies on a two-expert (LSTM-KF vs. GameFormer) pool. Table~\ref{tab:gating_comparison} summarises the full results, derived from our research log. This comprehensive comparison justifies our choice of the pairwise ranking-gate formulation as presented in the main text.

The analysis confirms that meta-features are essential (Methods 5--9) and that a Ranking-based formulation (Method 8) provides the best balance of mean FDE performance and oracle realisation. Methods 1--3, which rely solely on geometric features, fail to achieve meaningful oracle realization, validating our emphasis on model-internal signals.

\begin{table*}[!t]
\centering
\caption{Comprehensive comparison of all evaluated gating strategies (Two-Expert Pool: LSTM-KF vs. GameFormer). We report Final Displacement Error (FDE), Oracle Realisation Rate (ORR), 95th Percentile Tail Risk (P95 FDE), and the computational overhead (fraction of GameFormer (GMF) inferences required). Ranking-Gate (Method 8) achieves the best overall FDE and ORR, while the Risk-Gate (Method 7) is optimal for tail-risk (P95) mitigation.}
\label{tab:gating_comparison}
\resizebox{\textwidth}{!}{%
\begin{tabular}{@{}llclcccr@{}}
\toprule
\textbf{ID} & \textbf{Gating Strategy} & \textbf{Primary Features} & \textbf{FDE (m)}$\downarrow$ & \textbf{ORR}$\uparrow$ & \textbf{P95 FDE (m)}$\downarrow$ & \textbf{GMF Cost}$\downarrow$ & \textbf{Key Insight} \\
\midrule
\multicolumn{8}{l}{\textit{Baselines \& Theoretical Bound}} \\
-- & LSTM-KF (Baseline) & N/A & 7.7156 & -- & 24.81 & 0\% & Physics-based baseline \\
-- & GameFormer (Baseline) & N/A & 4.4548 & 0.0\% & 16.47 & 100\% & SOTA baseline \\
-- & Oracle (Upper Bound) & N/A & 3.6026 & 100.0\% & -- & -- & Theoretical best \\
\midrule
\multicolumn{8}{l}{\textit{Gating Experiments (Methods 1--10)}} \\
1 & Thresholding & Geometric & 4.4406 & 1.7\% & -- & \textless{}1\% & Geometric features insufficient \\
2 & MLP Classification & Geometric & \textasciitilde{}4.45 & \textasciitilde{}0\% & -- & \textless{}3\% & Failed, heavily biased to GMF \\
3 & MLP (Improved) & Geometric & \textgreater{}4.45 & \textless{}0\% & -- & 100\% & Failed, high GMF misclassification \\
4 & Regression (Predict $\Delta$FDE) & Meta-features & 4.4103 & -5.2\% & -- & 100\% & Failed (R² = -0.005) \\
5 & Meta-Gate (Classification) & Meta-features & 4.2953 & 18.7\% & -- & 100\% & Meta-features are effective \\
6 & Quantile Regression (Q75) & Meta-features & 4.3576 & \textasciitilde{}14\% & -- & 100\% & Systematically overestimates risk \\
7 & Risk-Based Gate (Q90) & Meta-features & 4.4154 & \textasciitilde{}15\% & \textbf{15.89} & 100\% & \textbf{Best for Tail Risk (P95)} \\
\textbf{8} & \textbf{Ranking-Gate (Ours)} & \textbf{Meta-features} & \textbf{4.2243} & \textbf{27.0\%} & -- & \textbf{100\%} & \textbf{Best Overall FDE/ORR} \\
9 & Two-Stage Gate & Meta-features & 4.2974 & 18.5\% & -- & 90.1\% & Best compute/performance trade-off \\
\bottomrule
\end{tabular}%
}
\end{table*}

\subsection{Appendix C: Supplementary Materials and Reproducibility}
\label{sec:supplementary}

Per thesis requirements, this section provides access to supplementary materials and original supporting documents that serve as direct evidence for the work completed in the research paper. These artifacts are intended for reference by the examiners and to support full reproducibility of our results.

\paragraph{Code Repository and Implementation.}
The complete source code for the experimental framework described in this paper is publicly available. This repository includes the implementation of the meta-feature extraction pipeline, the pairwise ranking-gate, the LLM supervisor module, and all evaluation and ablation scripts.

\begin{itemize}
    \item \textbf{GitHub Repository:} 
    
    \url{https://github.com/lbw1850151881-lang/trajectory-ranking-gate}
\end{itemize}

\paragraph{Experimental Records and Data.}
The complete experimental logs, raw results files, evaluation outputs, and research records (including those referenced in the paper) are archived as supporting documentation for the ``Substance of Research''. This evidence validates the experimental protocols and the results presented in Section~\ref{sec:results}.

\begin{itemize}
    \item \textbf{Google Drive Archive:} 
    
    \url{https://drive.google.com/drive/folders/1QcIQSsxMhPQiKipxWi20QsvVUMaOR-cV?usp=sharing}
\end{itemize}

\paragraph{Project Demonstration Page.}
A public-facing project page provides a high-level overview of the research, visualizations, and demonstrations.

\begin{itemize}
    \item \textbf{Project Page:} 
    
    \url{https://lbw1850151881-lang.github.io/trajectory-ranking-gate/}
\end{itemize}

\end{document}